# Optimization of Worker Scheduling at Logistics Depots Using Genetic Algorithms and Simulated Annealing

Haixin Wu[#]
Academy of Engineering
Peoples' Friendship University of Russia
Moscow, Russia
2328077194@qq.com

Jinxin Xu[#]
Department of Cox Business School
Southern Methodist University
Dallas, America
jensenjxx@gmail.com

Yu Cheng[#]
The Fu Foundation School of Engineering and Applied Science
Columbia University
New York, America
yucheng576@gmail.com

Liyang Wang
Olin Business School
Washington University in St. Louis
St. Louis, America
liyang.wang@wustl.edu

Xin Yang
College of Information Engineering
Sichuan agriculture university
Ya'an, China
3021265889@qq.com

Xintong Fu
China United Network Communications Group Co., Ltd.
Beijing, China
fuxt6@chinaunicom.cn

Yuelong Su*
School of Urban and Environmental Sciences
Central China Normal University
Wuhan, China
mersuyl@mails.ccnu.edu.cn

[#]These authors contributed equally to this work

*Abstract*—**This paper addresses the optimization of scheduling for workers at a logistics depot using a combination of genetic algorithm and simulated annealing algorithm. The efficient scheduling of permanent and temporary workers is crucial for optimizing the efficiency of the logistics depot while minimizing labor usage. The study begins by establishing a 0-1 integer linear programming model, with decision variables determining the scheduling of permanent and temporary workers for each time slot on a given day. The objective function aims to minimize person-days, while constraints ensure fulfillment of hourly labor requirements, limit workers to one time slot per day, cap consecutive working days for permanent workers, and maintain non-negativity and integer constraints. The model is then solved using genetic algorithms and simulated annealing. Results indicate that, for this problem, genetic algorithms outperform simulated annealing in terms of solution quality. The optimal solution reveals a minimum of 29857 person-days.**

## I. Introduction

Improving the efficiency of sortation center management has a direct impact on the fulfillment efficiency and operational costs of the entire logistics network. Staff management in sortation centers is a key challenge. Staffing needs to be adjusted according to the forecasted shipment volume to ensure a sufficient workforce to handle the flow of goods during peak hours while avoiding the wastage of excess manpower during low-demand times. Staff scheduling based on effective solution algorithms becomes one of the key strategies to improve the efficiency of the sorting center. By reasonably allocating regular and temporary workers, the sorting speed and accuracy can be improved, thus reducing the overall logistics cost and improving customer satisfaction.

Currently, some scholars have already launched related research. zhe Sun et al [1] established a new model for warehouse human resource scheduling and proposed an adaptive quantum differential evolution (AQDE) algorithm, inspired by quantum computing, to solve the optimization complexity problem. The method demonstrated superior global convergence accuracy and speed compared to other DE variants and evolutionary algorithms. Ning Lei [2] proposed an intelligent distribution model based on the Internet of Things. The model not only optimizes the distribution process but also proposes an efficient distribution strategy when faced with large amounts of data. This technique mainly uses information interaction technology in the Internet of Things, which can ensure the fastest average distribution speed, the shortest average transport distance, and the shortest time consumed in the logistics transmission process. In the decision-making process of this intelligent distribution model, some intelligent distribution models controlled by multiple indicators are first built. During the experiments, a logistics-aware system was constructed and a heuristic algorithm was used to solve the crating problem. P. Milička et al [3] proposed a two-layer optimization model that embeds a nested internal optimization problem (i.e., the project manager's decision-making problem) as a constraint on the external optimization problem (i.e., the team leader's decision-making problem). The algorithm is a mathematical programming approach that generates additional inert constraints through feasibility callbacks so that the team leader problem has to comply with the requirements formulated in the project manager's problem. Renata Mansini et al [4] investigated a complex multi-objective staff scheduling problem driven by a real case. A multi-objective mixed integer linear programming model of the problem is given. The constraints are categorized into mandatory and optional constraints. This work introduces a solution methodology, called PRIMP (Prioritise and Improve), which enforces the constraints to be satisfied by employing an additional objective function. Ana Batista et al [5] proposed a mixed integer linear programming model that considers a microscopic Markov chain approach for determining the probability of infection in a contact network that mimics employee interactions. The model determines, for a given planning period, the optimal staffing mix to maximize occupancy while minimizing the risk of infection in the presence of a detection protocol.

The efficient scheduling of workers in logistics depots, comprising both permanent and temporary staff, is a critical aspect of operational optimization. This scheduling directly impacts the overall efficiency of the depot's operations and is pivotal in managing labor resources effectively. To address this, the study formulates a 0-1 integer linear programming model to optimize the scheduling of workers. The model's decision variables dictate the presence of permanent and temporary workers across different shifts on each day, while the objective function aims to minimize person-days. Several constraints are imposed to ensure that hourly labor requirements are met, workers are scheduled for only one-time slot per day, and permanent workers' consecutive working days do not exceed seven. Additionally, the model is subject to non-negativity and integer constraints. To solve this complex optimization problem, genetic algorithms and simulated annealing are employed. This paper presents a comparative analysis of the performance of these algorithms and provides insights into their respective strengths and weaknesses in solving the given scheduling optimization problem.

## II. DESCRIPTION OF THE PROBLEM

The e-commerce logistics network consists of several links in order fulfillment. Among them, the sorting center, as an intermediate link in the network, needs to sort parcels according to different flow directions and send them to the next site, so that the parcels finally reach the hands of consumers. The improvement of sorting center management efficiency plays a very important role in the overall network performance efficiency and operating costs.

Personnel scheduling based on the sorting center volume forecast is the next important problem to be solved, the sorting center personnel contains two kinds of regular workers and temporary workers: regular workers are the personnel employed for a long time at the site, with high working efficiency: temporary workers are the personnel recruited temporarily according to the volume of goods, and can be added or removed at will every day, but with relatively low working efficiency and high hiring costs. Reasonable arrangement of personnel aims to complete the work as far as possible to reduce staff costs. For the logistics network considered in this paper, the staffing schedule and hourly manpower efficiency indicators are as follows.

1) For all sorting centers, each day is divided into 6 shifts, respectively: 00:00-08:00, 05:00-13:00, 08:00-16:00, 12:00-20:00, 14:00-22:00, 16:00-24:00, and each person (regular or temporary) can only be on duty for one shift per day.

2) The hourly efficiency index is the number of parcels sorted per person per hour (the number of parcels i.e. the number of goods), the maximum hourly efficiency of regular workers is 25 parcels/hour, and the maximum hourly efficiency of temporary workers is 20 parcels/hour.

## III. MATHEMATICAL MODEL

A mathematical model is developed based on the problem described in Section II.

A 0-1 integer linear programming model is developed as shown below:

Decision variable:

$x_{d,t,s}$: decision variable (0 or 1) for whether a regular laborer, time period t, works the sth shift on day d or not

$y_{d,s}$: decision variable (0 or 1) for whether a temporary worker works the sth shift on day d or not

The remaining variables are as follows:

$C_{d,s}$: daily cargo demand per shift.

$P$: Hourly wage for regular workers.

$T$: Hourly wage for temporary worker $s$.

$H_{d,t}$: the number of human efficiencies in the t-th time period of the dth day.

$A_s$: Attendance rate of regular workers.

The objective function is as follows:

$$Minimize \sum_{d=1}^{30}\sum_{t=1}^{6}\sum_{s=1}^{200} x_{d,t,s} + \sum_{d=1}^{30}\sum_{t=1}^{6} y_{d,t} \quad (1.)$$

Constraints:

1) Satisfy the manpower efficiency demand per hour:- This constraint ensures that at any given time period, the

number of workers scheduled satisfies the manpower efficiency demand for that time period. -For each time period t on each day d, the following conditions are satisfied.

$$\sum_{s=1}^{200} x_{d,t,s} \cdot P + y_{d,t} \cdot T \geq C_{d,t} \qquad (2.)$$

where $x_{d,t,s}$ denotes whether regular worker s is working at time period t on day d, $y_{d,t}$ denotes whether temporary workers are working at time period t on day d, and $C_{d,t}$ is the demand for goods at time period t on day d.

(2) Each worker works a maximum of one time period in a day:

This constraint ensures that each worker works at the most one-time slot in a day.

For each regular worker s and each day d, the following conditions are satisfied:

$$\sum_{t=1}^{6} x_{d,t,s} \leq 1 \qquad (3.)$$

(3) The number of consecutive days of attendance per regular worker cannot exceed seven:

This constraint limits the number of consecutive days of attendance to no more than seven for each regular worker.

For each regular worker s and each day d, the following conditions are satisfied:

$$\sum_{d'=d}^{d+6} \sum_{t=1}^{6} x_{d',t,s} \leq 7 \qquad (4.)$$

(4) Non-negative and integer constraints:

This constraint ensures that the decision variables $x_{d,t,s}$ and $y_{d,t}$ are non-negative integers.

For each formal worker s, each day d and each time period t, the following conditions are satisfied:

$$x_{d,t,s}, y_{d,t} \geq 0, \text{ and } x_{d,t,s}, y_{d,t} \in \{0,1\}1 \qquad (5.)$$

These constraints ensure the rationality of the work arrangement, guarantee the satisfaction of the production demand, and have some limitations on the work arrangement and attendance of the regular workers.

In summary, the linear programming model established in this paper can be formulated as:

Objective function:

$$Minimize \sum_{d=1}^{30} \sum_{t=1}^{6} \sum_{s=1}^{200} x_{d,t,s} + \sum_{d=1}^{30} \sum_{t=1}^{6} y_{d,t} \qquad (6.)$$

Constraints:

$$\sum_{s=1}^{200} x_{d,t,s} \cdot P + y_{d,t} \cdot T$$
$$\geq C_{d,t} \sum_{k=0}^{7} \sum_{T=1}^{6} x_{i,d+k,T} \qquad (7.)$$
$$\leq 7$$

$$\sum_{t=1}^{6} x_{d,t,s} \leq 1 \qquad (8.)$$

$$\sum_{d'=d}^{d+6} \sum_{t=1}^{6} x_{d',t,s} \leq 7 \qquad (9.)$$

$$x_{d,t,s}, y_{d,t} \geq 0 \qquad (10.)$$

$$x_{d,t,s}, y_{d,t} \in \{0,1\} \qquad (11.)$$

IV. MODEL SOLVING

In this paper, due to the large solution space under consideration, the model is solved using an intelligent optimization algorithm as the computation time is longer if a solver is used for the solution.

Intelligent optimization algorithms have greater flexibility and applicability in mathematical model solving compared to traditional solvers. Intelligent optimization algorithms are able to deal with more complex problems, including non-linear, non-convex, and even problems without explicit expressions. This means that intelligent optimization algorithms are able to provide more comprehensive solutions when faced with actual complex mathematical models.

In addition, intelligent optimization algorithms typically have a wider range of parameters and strategies that can be adapted to a particular problem. This flexibility means that algorithms can be customized to better suit different types of mathematical models. In contrast, some traditional solvers may have limited ability to handle complex problems, especially in solving nonlinear and nonconvex problems.

A. *Genetic Algorithm-Based Model Solving*

Genetic algorithm, as one of the most commonly used intelligent optimization algorithms, has been developed and used by many scholars for its excellent exploration and exploration performance. Genetic algorithm imitates the evolutionary process of the population in biology when performing optimization, and explores the solution space under the premise of retaining the excellent individuals and gene fragments in the population through the two operators of crossover and mutation. The crossover operator is used to complete the exchange and dissemination of good gene segments among individuals in the population, and the mutation operator is used to complete the exploration of the solution space in order to avoid falling into the local optimum. The framework of the genetic algorithm is shown in Fig. 1.

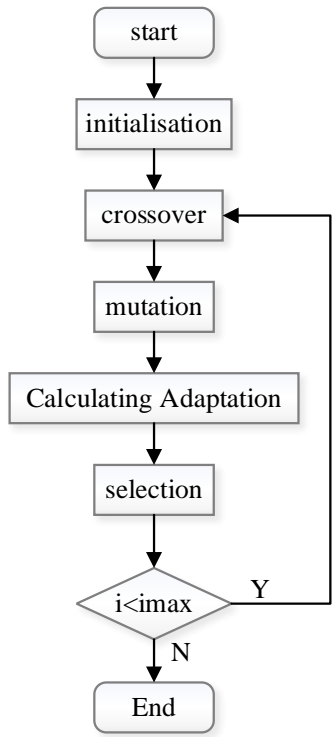

Fig. 1. Genetic Algorithm Process

As can be seen in Figure 1, the genetic algorithm contains five main parts.

(1) Initialisation. In this stage, the chromosomes need to be encoded to generate an initial population. The genetic algorithm does not act directly on the solution space of the real problem but needs to use the encoding to describe the real problem first, and then evolve the solution based on the encoding. The initial population, often contains multiple individuals, each of which consists of a set of chromosomes representing a solution.

(2) Crossover. This is an important way of spreading good genes in genetic algorithms. In this step, two individuals swap chromosome segments with a certain probability.

(3) Mutation. Mutation is an important way for genetic algorithms to jump out of the local optimum and maintain the diversity of solutions in the population. In the mutation operation, a certain gene segment in the chromosome is generally changed with a certain probability to achieve the purpose of exploring the solution space to jump out of the local optimum.

(4) Calculation of fitness. In this step, it is necessary to decode the chromosome, and decoding the chromosome is not the inverse process of encoding in the workshop scheduling problem. There are often remaining constraints in decoding that are complementary, such as machine availability constraints, material availability constraints, and so on. The fitness also represents the objective value of the problem being solved.

(5) Selection. Based on the objective value obtained, the performance of a solution can be judged to a certain extent. At this time, some individuals can be selected for the next iteration, in this process, the general idea is: to retain the excellent individuals, and eliminate the poor individuals, this idea is also known as the greedy selection strategy. However, in practice, the selection is often relaxed to a certain extent, in order to avoid mistakenly deleting individuals carrying excellent genes or the population into the local optimal situation. As a result, the mechanism of selection can be extended to roulette selection, bidding race selection and so on. The essence of all of them is the relaxation relative to the greedy selection strategy.

For the problem solved in this paper, three-time dimensions need to be considered. One is the date, the second is the time of day, and the third is the hour. If vector or matrix coding is used it will lead to inefficient problem solving. Therefore, this paper designs a multi-chromosome coding scheme to represent the solution. As shown in Fig. 2

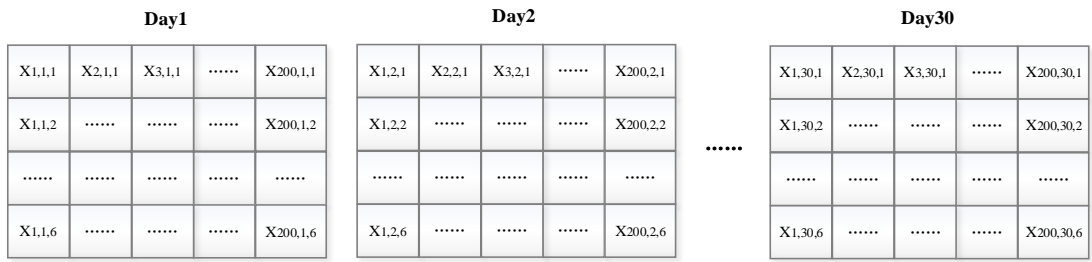

Fig. 2. encoding scheme

In this paper, the temporary labor arrangement is derived by decoding. This is done in the following way: the number of temporary workers = the difference in the required distribution volume. Arranging the temporary workers in this way can reduce the complexity of the infeasible solution repair.

In the scheme devised in this paper, crossover is performed on any 15-day chromosome of an individual if it is selected for crossover. The crossover method is shown in Fig. 3.

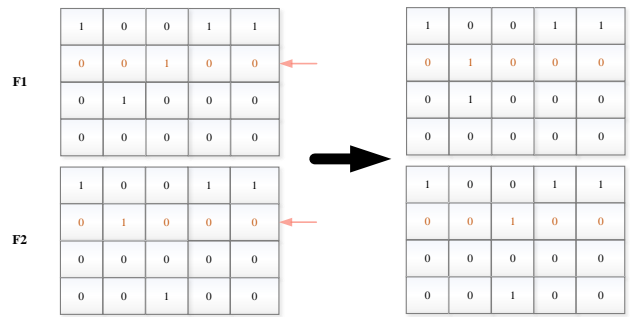

Fig. 3. crossover strategy

In this paper, the following constraints need to be considered to fix the infeasible solution:

1) The distribution of goods per hour needs to be completed;

2) Each worker can only work at one-time slot in a day;

3) Each regular worker must not work for more than 7 consecutive days;

4) the attendance of any one regular worker cannot exceed 85%.

For restriction 1, this paper designs the decoding scheme in order to circumvent the problem, or the difference in the cargo volume distribution will be made up by temporary workers in the decoding phase.

For the remaining three restrictions, the fixing strategy is shown below:

For restriction 3), iterate through each regular worker and assign 0 to the 8th consecutive working day if they work more than 7 consecutive days;

For restriction 2), iterate over each regular worker, and if a work period is exceeded on a particular day for a regular worker, leave one work period at random and clear 0 to the rest;

(c) For limitation 4), for each regular worker, if attendance exceeds 85 percent, randomly delete attendance on a given day until attendance is no higher than 85 percent.

*B. simulated annealing algorithm*

Simulated Annealing is a probabilistic optimization algorithm that draws inspiration from the physical process of annealing in metallurgy. The algorithm aims to find the global optimum in a large solution space by allowing the algorithm to accept worse solutions with a certain probability, which decreases over time. This enables the algorithm to escape local optima and explore a wider solution space.

The steps of the algorithm are shown below:

Initialization: Simulated Annealing begins with an initial solution, often randomly generated or based on some heuristic method. Additionally, a starting temperature and a cooling schedule are defined. The cooling schedule determines how the temperature decreases over time.

Neighborhood Search: At each iteration, the algorithm explores a neighboring solution. This can be achieved through various methods such as small random perturbations to the current solution.

Acceptance Criterion: The algorithm compares the neighboring solution with the current solution. If the neighboring solution is better, it is always accepted. If the neighboring solution is worse, it may still be accepted with a certain probability, which depends on the current temperature and the degree of worsening. This probability allows the algorithm to escape local optima early in the search.

Cooling: As the algorithm progresses, the temperature is reduced according to the cooling schedule. The temperature reduction decreases the probability of accepting worse solutions, gradually shifting the algorithm's focus towards exploitation rather than exploration.

The simulated annealing algorithm in this paper is encoded in a similar way to the genetic algorithm, so it will not be repeated.

*C. Algorithm results*

The convergence plots obtained by the two algorithms are shown in Fig. 4.

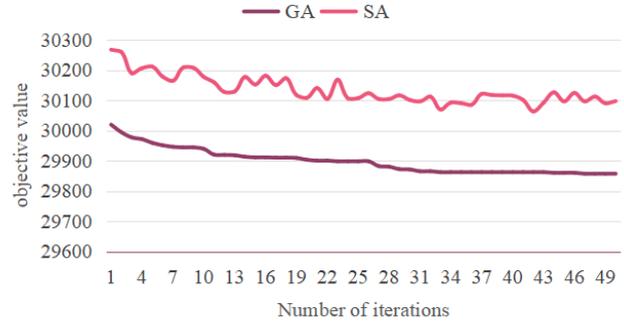

Fig. 4. convergence curve

The convergence curve in Fig. 4 shows that the genetic algorithm used is able to solve the optimization model developed in this paper better compared to the simulated annealing algorithm. This is due to the fact that the genetic algorithm as a swarm intelligence algorithm is able to explore the solution space better. At the same time, the genetic algorithm can explore the solution space through two measures: crossover and mutation, so the simulated annealing algorithm is prematurely compared to the genetic algorithm, and the solution obtained is slightly inferior to the genetic algorithm.

V. CONCLUSION

In this study, the optimization of worker scheduling at a logistics depot was examined using a combined approach of genetic algorithms and simulated annealing. The results of the analysis demonstrate that genetic algorithms outperform simulated annealing in terms of solution quality. The optimal solution reveals a minimum of 29857 person-days required for the given scheduling problem. This research underscores the significance of employing advanced optimization techniques in addressing complex real-world scheduling challenges and provides valuable insights into the comparative performance of genetic algorithms and simulated annealing in the context of worker scheduling optimization.